# Efficient Stepwise Selection in Decomposable Models

**Amol Deshpande***
Computer Science Department
University of California
Berkeley, CA 94720
amol@cs.berkeley.edu

**Minos Garofalakis**
Bell Laboratories
600 Mountain Avenue
Murray Hill, NJ 07974
minos@research.bell-labs.com

**Michael I. Jordan**
Computer Science & Statistics
University of California
Berkeley, CA 94720
jordan@cs.berkeley.edu

## Abstract

In this paper, we present an efficient algorithm for performing stepwise selection in the class of decomposable models. We focus on the *forward selection* procedure, but we also discuss how *backward selection* and the combination of the two can be performed efficiently. The main contributions of this paper are (1) a simple characterization for the edges that can be added to a decomposable model while retaining its decomposability and (2) an efficient algorithm for enumerating all such edges for a given decomposable model in $O(n^2)$ time, where $n$ is the number of variables in the model.

We also analyze the complexity of the overall stepwise selection procedure (which includes the complexity of enumerating eligible edges as well as the complexity of deciding how to "progress"). We use the KL divergence of the model from the saturated model as our metric, but the results we present here extend to many other metrics as well.

## 1 Introduction

Undirected graphical models have become increasingly popular in areas such as information retrieval, statistical natural language processing, and vision, where they are often referred to as *maximum entropy* or *Gibbs* models, and are viewed as having various representational and statistical advantages. New tools for model selection and parameter estimation are being developed by researchers in these areas [PPL97, Hin99, ZWM97]. General undirected models, however, have some serious disadvantages, in particular they require an invocation of Iterative Proportional Fitting (or related iterative algorithms) to find maximum likelihood estimates, even in the case of fully-observed graphs. As the inner loop of more general parameter estimation or model selection procedures (e.g., the M step of an EM algorithm), these iterative algorithms can impose serious bottlenecks.

Decomposable models are a restricted family of undirected graphical models that have a number of appealing features: (1) maximum likelihood estimates can be calculated analytically from marginal probabilities, obviating the need for Iterative Proportional Fitting, (2) closed form expressions for test statistics can be found, and (3) there are several useful links to directed models (every decomposable model has a representation as either an undirected or a directed model), inference algorithms (decomposable models are equivalent to triangulated graphs), and acyclic database schemes [BFMY83]. Decomposable models would therefore seem to provide a usefully constrained representation in which model selection and parameter estimation methods can be deployed in large-scale problems. Moreover, mixtures of decomposable models provide a natural upgrade path if the representational strictures of decomposable models are considered too severe.

Although decomposable models are a subclass of undirected graphical models, the problem of finding the optimal decomposable model for a given data sample is known to be intractable and heuristic search techniques are generally used [Edw95]. Most procedures are based on some combination of (i) forward selection, in which we start with a small model and add edges as long as an appropriate score function increases [Hec98], and (ii) backward selection, where starting with a larger model, edges are deleted from the model. Since the intervening models encountered in the search must also be decomposable, care must be taken such that deletion or addition of edges does not result in a non-decomposable model. Backward selection procedures for decomposable models are well known in the literature [Wer76, Lau96], but efficient forward selection procedures have not yet been developed. One of the goals of the current paper is to fill this gap.

---

*Part of the work was done while the author was visiting Bell Laboratories.

This paper is a theoretical paper that makes two main contributions. First, we provide a simple characterization of the edges that can be added to a decomposable model (or, equivalently, the chordal graph corresponding to the model) while resulting in another decomposable model. Second, based on this characterization we present an efficient algorithm for enumerating all such edges for the current model in $O(n^2)$ time, where $n$ is the number of attributes. We provide a careful analysis of the running time complexity of the overall forward selection procedure, including the time taken for choosing which of the eligible edges to add to the current model. We use the minimization of KL divergence as our metric, but the results we present can be extended to any other locally computable metric (e.g., [GG99]).

Though our main focus is the new forward selection procedure, we also show that the algorithms are easily extended to backward selection or to a combination of forward and backward selection. The techniques and data structures we propose also naturally extend to the problem of finding strongly decomposable models in mixed graphs.

The remainder of the paper is organized as follows. In Section 2, we derive a simple characterization for the edges that can be added to a chordal graph while maintaining its chordality. In Section 3, we describe the data structure that we use for finding such edges efficiently and discuss how it is maintained in the presence of additions to the underlying model graph. In Section 4, we analyze the overall complexity of the stepwise selection procedures for the KL divergence metric. We briefly discuss how the data structures can be extended for doing backward selection as might be required for a procedure that alternates between forward and backward selection in Section 5. In Section 6, we discuss how these algorithms can be extended for the case of *mixed graphs* and strong decomposability. We conclude with Section 7.

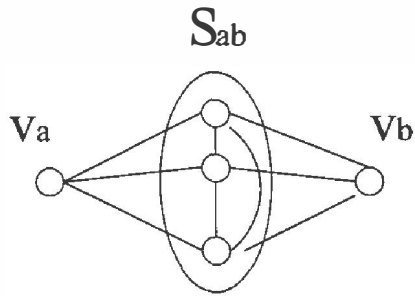

Figure 1: Structure of the subgraph induced by $v_a, v_b$ and $S_{ab}$

## 2 Characterizing Edges Eligible for Stepwise Selection

There is a classical characterization of the edges that can be deleted from a chordal graph[1] such that the resulting graph remains chordal:

**Theorem 2.1** [Wer76, Lau96] Given a chordal graph $\mathcal{G}$, an edge can be deleted from the graph while maintaining its chordality iff the edge belongs to exactly one of the maximal cliques of the graph.

To complement this result, we propose the following characterization of the edges that can *added* to a chordal graph without violating its chordality (the proof is given in Appendix A):

**Theorem 2.2** Given a chordal graph $\mathcal{G} = (V, E)$, an edge $(v_a, v_b) \notin E$ can be added to the graph while maintaining its chordality iff it satisfies the following properties:

1. There exists a subset of nodes $S_{ab} \subseteq V - \{v_a, v_b\}$, such that $(v_a, u), (v_b, u) \in E, \forall u \in S_{ab}$; i.e., $v_a$ and $v_b$ are connected to all vertices in $S_{ab}$;

2. The set $S_{ab}$ is the *minimal separator* for $v_a$ and $v_b$ in $\mathcal{G}$ (note that, since $\mathcal{G}$ is chordal, this implies that $S_{ab}$ is a clique); i.e., removing the nodes in $S_{ab}$ and all their incident edges from $\mathcal{G}$ separates $v_a$ and $v_b$ and no proper subset of $S_{ab}$ has this property.

## 3 Enumerating Eligible Edges for Forward Selection

In this section, we describe how to enumerate all edges that can be added to the current chordal graph while maintaining chordality. For this purpose, we maintain two auxiliary data structures (i) a *clique graph* [GHP95] corresponding to the current chordal graph, (ii) a boolean $n \times n$ matrix, $\mathcal{E}_\mathcal{M}^f$, indexed by the attributes of data which maintains information about whether a pair of nodes is eligible for addition. Clearly, using $\mathcal{E}_\mathcal{M}^f$, we can enumerate the edges eligible for forward selection in $O(n^2)$ time, where $n$ is the number of attributes. The clique graph is required to update $\mathcal{E}_\mathcal{M}^f$ in $O(n^2)$ time when an edge is added to the underlying graph, as we will see later in the section.

---
[1] For simplicity, we use the terms decomposable model and chordal graph interchangeably.

## 3.1 Clique Graph

### 3.1.1 Definition and Properties

**Definition:** A *clique graph* of a chordal graph $\mathcal{G} = (V, E)$, denoted by $CG_\mathcal{G}$, has the maximal cliques of the chordal graph as its vertices, and has the property that given any two maximal cliques, $C_1$ and $C_2$, there is an edge between these two cliques iff $C_1 \cap C_2$ separates the node sets $C_1 \setminus C_1 \cap C_2$ and $C_2 \setminus C_1 \cap C_2$.

Note that this usage of the term "clique graph" is non-standard; we are following the terminology of [GHP95].[2]

We will note some properties of clique graphs.

**Lemma 3.1** Two nodes $v_a, v_b \in V$ with $(v_a, v_b) \notin E$ satisfy the forward selection characterization (Section 2.2) iff there exists an edge $(C_1, C_2)$ in $CG_\mathcal{G}$ such that $v_a \in C_1$ and $v_b \in C_2$.

**Lemma 3.2** [GHP95] A maximum spanning tree of a clique graph of a chordal graph, where the weight of an edge is the size of the intersection of the two cliques it joins, is a *junction tree* of the chordal graph.

**Lemma 3.3** The number of nodes in a clique graph of a chordal graph $\mathcal{G} = (V, E)$ is at most $|V|$.

## 3.2 Updating the Clique Graph

After all possible edges are enumerated and one edge is chosen based on a model selection criterion, we need to update the clique graph to reflect the addition of this new edge to the underlying chordal graph. In this section, we show how this can be done in $O(n^2)$ time.

---

[2] Note also that the clique graph of a chordal graph is equivalent to an Almond Junction Tree [AK93, JJ94] of the graph; both of these structures can be seen as compact representations of all possible junction trees of a chordal graph.

### 3.2.1 Update Algorithm

Let $\mathcal{G} = (V, E)$ be the original chordal graph and let $G_\mathcal{G}$ be the clique graph. Let the new edge that is added be $(v_a, v_b)$. Also, let $(C_a, C_b)$ be the corresponding edge in the clique graph (the edge from which this pair of nodes was obtained) and let $S_{ab} = C_a \cap C_b$. Finally, let $\mathcal{G}'$ and $CG_{\mathcal{G}'}$ be the new model and clique graphs.

Addition of the edge $(v_a, v_b)$ creates a new maximum clique $C_{ab} = S_{ab} + v_a + v_b$ as shown in Figure 2. It is possible that $C_a \subset C_{ab}$ (or $C_b \subset C_{ab}$), in which case, $C_a$ ($C_b$) will not be a maximal clique in the new chordal graph. We assume for now that this does not happen. In that case, adding the edge $(v_a, v_b)$ to $\mathcal{G}$ results in a partial clique graph structure as shown in Figure 2. Note that the new clique will be connected to $C_a$ and $C_b$ in $CG_{\mathcal{G}'}$.

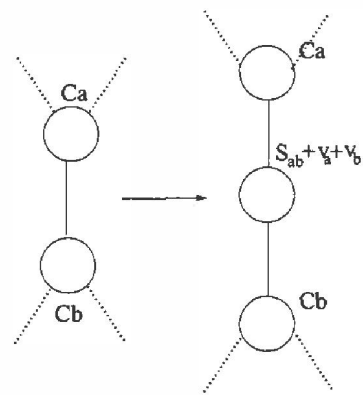

Figure 2: Partial Structure of the Clique Graph after adding the edge $(v_a, v_b)$

After this node has been created and added to the clique graph, the update algorithm has to delete those edges from the clique graph which do not satisfy the clique graph property (Section 3.1.1). It is easy to see that no new edges between already existing maximal cliques in the clique graph will have to be added. Given this, the update algorithm is as follows (see Appendix B for the correctness proof):

1. Let $\mathcal{G}'' = \mathcal{G} - S_{ab}$. Find all nodes that are connected to $v_a$ and to $v_b$ in $\mathcal{G}''$ and maintain this information in two arrays indexed by the vertices of $\mathcal{G}$.

2. **Deciding whether to keep an edge** $(C_1, C_2) \in CG_\mathcal{G}$ **in** $CG_{\mathcal{G}'}$: Let $S_{12} = C_1 \cap C_2$. If $S_{12} \neq S_{ab}$, then keep this edge. Otherwise, consider the graph $\mathcal{G} - S_{12} (= \mathcal{G}'')$. If $v_a$ is connected to $C_1 \setminus S_{12}$ in this graph and $v_b$ is connected to $C_2 \setminus S_{12}$ or *vice* *versa*, do not keep this edge in $CG_{\mathcal{G}'}$. Otherwise, keep it. This check can be performed in $O(1)$ time using the arrays constructed in Step 1.

3. **Adding edges involving $C_{ab}$:**

   1. For every maximal clique $C'$ such that $(C', C_a) \in CG_{\mathcal{G}'}$ (after execution of the above step), add an edge $(C', C_{ab})$ to $CG_{\mathcal{G}'}$ if $(C' \cap C_{ab}) \subset S_{ab} + v_a$.
   2. For every maximal clique $C'$ such that $C' \cap C_{ab} = S_{ab} + v_a$, explicitly check if this edge should be added to the clique graph. This requires checking whether $v_b$ is separated from $C' \setminus S_{ab} + v_a$ by $S_{ab} + v_a$, which can be done by looking up the arrays computed in Step 1.
   3. Repeat the above two steps for $C_b$ and $v_b$ instead of $C_a$ and $v_a$.

4. Remove $C_a$ ($C_b$) if it is contained in $C_{ab}$.

---

**Input:** $\mathcal{G} = (V, E)$, $CG_{\mathcal{G}}$, current model & clique graphs
**Output:** $\mathcal{G}' = (V, E')$, $CG_{\mathcal{G}'}$, new model & clique graphs
**begin**
1. using $\mathcal{E}^f_{\mathcal{M}}$, choose which of the edges to add to $\mathcal{G}$
2. let $(v_a, v_b)$ be the new edge to be added, let $(C_a, C_b) \in CG_E$ be the corresponding edge in $CG_{\mathcal{G}}$ and let $S_{ab} = C_a \cap C_b$.
3. $\mathcal{G}'' \leftarrow \mathcal{G} - S_{ab}$. Find all nodes connected to $v_a$ and to $v_b$ in $\mathcal{G}''$ and maintain this information in two arrays indexed by the vertices of $\mathcal{G}$.
4. $\mathcal{G}' \leftarrow (V, E + (v_a, v_b))$.
5. $CG_{\mathcal{G}'} \leftarrow CG_{\mathcal{G}}$.
6. delete $(C_a, C_b)$ from $CG_{\mathcal{G}'}$, add a new node $C_{ab} = S_{ab} + v_a + v_b$ and add edges $(C_a, C_{ab})$ and $(C_b, C_{ab})$.
7. **for all** edges $(C_1, C_2) \in CG_E$ **do**
8.    **if** $S_{12} = C_1 \cap C_2$ equal to $S_{ab}$ **then**
9.       **if** $v_a$ is connected to $C_1 \setminus S_{ab}$ and $v_b$ is connected to $C_2 \setminus S_{ab}$ or *viceversa* **then**
10.          delete the edge $(C_1, C_2)$ from $CG_{\mathcal{G}'}$.
11. **for all** $C' \in CG_V$ s.t. $(C', C_a) \in CG_{\mathcal{G}'}$ **do**
12.    $S' \leftarrow C' \cap C_a$.
13.    **if** $S' \subset S_{ab} + v_a$ **then**
14.       add the edge $(C_{ab}, C')$ to $CG_{\mathcal{G}'}$
15. **for all** $C' \in CG_V$ s.t. $C' \cap C_{ab} = S_{ab} + v_a$ **do**
16.    check if $S_{ab} + v_a$ separates $v_b$ from $C' \setminus (C' \cap C_{ab})$
17.    if yes, add the edge $(C_{ab}, C')$ to $CG_{\mathcal{G}'}$
18. repeat Steps 11-17 for $C_b$ and $v_b$ instead of $C_a$ and $v_a$.
19. remove $C_a$ (similarly, $C_b$) from $CG_{\mathcal{G}'}$ if $C_a \subset C_{ab}$.
20. update $\mathcal{E}^f_{\mathcal{M}}$ as described in Section 3.3
**end**

Figure 3: Complete Forward Selection Algorithm

---

### 3.2.2 Complexity Analysis

It is easy to see that Steps 1 and 3 can be performed in $O(n^2)$ time, but comparison of two sets to check if they are equal is $O(n)$ in general. As such, the complexity of Step 2 might seem to be $O(n^3)$ [3]. To perform this step efficiently, we maintain a *binary index* on the minimal separators of the current model graph. A *leaf* of this index contains a list of pointers to all the edges that have associated with them the minimal separator corresponding to the leaf. Now given $S_{ab}$, we can find all the edges with separator equal to $S_{ab}$ in time $O(n)$ (since the height of the index is $n$). As we mentioned in Step 2, we only need to check such edges since the rest of the edges remain untouched. The deletions, if required, can be performed in $O(1)$ time per deletion. Thus, given this data structure, Step 2 can be performed in time $O(n^2)$ (actually, in order $O(n + n')$ time, where $n'$ is the number of edges in $CG_{\mathcal{G}}$ with separator $S_{ab}$). Insertions in this index can be done in $O(n)$ time, by adding a pointer to the new edge at the front of the list of pointers. Since there can be at most $n$ insertions to this index in Step 3, the complexity of Step 3 remains $O(n^2)$.

### 3.3 Updating $\mathcal{E}^f_{\mathcal{M}}$

$\mathcal{E}^f_{\mathcal{M}}$ is updated as follows :

1. Set the entry $(v_a, v_b)$ to false, since this edge is not eligible for addition any more.

2. Since some edges are deleted from the clique graph, the corresponding pairs of nodes that might have been eligible for addition before may not be eligible any more. Noting that all such deleted clique graph edges must have the separator $S_{ab}$, we set the entry corresponding to pairs $(v_x, v_y)$ such that $v_x$ is connected to $v_a$ and $v_y$ is connected to $v_b$ in $\mathcal{G}'' = G - S_{ab}$, to false. This can be done by using the arrays computed during the clique graph update algorithm in $O(n^2)$ time.

3. Some edges of the form $(v_x, v_a)$ or $(v_x, v_b)$ may be now eligible for addition. To set the corresponding entries to true, for every edge that is added to the clique graph $(C', C_{ab})$, set to true the entries corresponding to $(v_x, v_a), v_x \in C', v_x, v_a \notin C' \cap C_{ab}$. Similarly for $v_b$ (please refer to Theorem 4.3 for the proof that these include all the edges newly

---

[3] In many cases, model search is performed among models with a restricted maximum clique size. In that case, set comparisons take constant time and the data structure we describe next is not required.

eligible for addition.).

Thus, $\mathcal{E}_{\mathcal{M}}^{f}$ can also be updated in $O(n^2)$ time.

## 4 Complexity of Overall Forward Selection Procedure

The overall forward selection procedure has two parts:

1. Enumerating all edges that can be added to the current model.
2. Deciding which of the eligible edges to add to the model for "forward progress".

Thus far, we have presented an $O(n^2)$ solution to the first problem, giving us an overall complexity of $O(n^2 k)$, if a total of $k$ edges are added starting with the null model. In this section, we analyze the complexity of the second step with the aim of minimizing KL divergence from the saturated model, which is equivalent to maximum likelihood estimation.

**Theorem 4.1** [Mal91] Minimizing the KL divergence of a given model $\mathcal{M}$ from the saturated model is equivalent to minimizing the entropy of the model $H(\mathcal{M})$ which is defined as:

$$H(\mathcal{M}) = \Sigma_{C \in \mathcal{C}} H(C) - \Sigma_{S \in \mathcal{S}} H(S)$$

where $\mathcal{C}$ is the set of maximum cliques of the model, and $\mathcal{S}$ is the set of separators for a junction tree of the model. The entropy of a subset of attributes, $A = (A_1, \ldots, A_k)$, is defined to be:

$$H(A) = \Sigma_{a \in D(A_1) \times \ldots \times D(A_k)} \frac{f(a)}{N} \log \frac{f(a)}{N}$$

where $D(A_i)$ denotes the domain of attribute $A_i$, $f(a)$ denotes the number of data items which contain $a$ and $N$ is the total number of data points.

Thus, the only information we need from a decomposable model, to compute its divergence from the saturated model, is a list of its maximal cliques and a list of separators in some junction tree for the model. Many metrics for graphical models have a similar property and this property can be used to avoid recomputation of a junction tree for the new model obtained after an incremental change to the current model:

**Theorem 4.2** If two decomposable models $\mathcal{M} \subset \mathcal{M}'$ differ only in one edge $(v_a, v_b)$, (i.e., $(v_a, v_b) \notin \mathcal{M}$ and $(v_a, v_b) \in \mathcal{M}'$), then the maximal cliques and the minimal separators in junction trees corresponding to the two models, $J_{\mathcal{M}} = (J_C, J_S)$ and $J_{\mathcal{M}'} = (J'_C, J'_S)$, where $J_C, J'_C$ are the lists of maximal cliques and $J_S, J'_S$ are the lists of separators, differ as follows:

1. If $C_a \not\subset C_{ab}$ and $C_b \not\subset C_{ab}$, then $J'_C = J_C + C_{ab}$ and $J'_S = J_S + C_{ab} \cap C_a + C_{ab} \cap C_b - S_{ab}$.
2. If $C_a \subset C_{ab}$ and $C_b \not\subset C_{ab}$, then $J'_C = J_C + C_{ab} - C_a$ and $J'_S = J_S + C_{ab} \cap C_b - S_{ab}$.
3. If $C_a \not\subset C_{ab}$ and $C_b \subset C_{ab}$, then $J'_C = J_C + C_{ab} - C_b$ and $J'_S = J_S + C_{ab} \cap C_a - S_{ab}$.
4. If $C_a \subset C_{ab}$ and $C_b \subset C_{ab}$, then $J'_C = J_C + C_{ab} - C_b - C_a$ and $J'_S = J_S - S_{ab}$.

This immediately gives us the following corollary:

**Corollary 4.1** If two decomposable models $\mathcal{M} \subset \mathcal{M}'$ differ only in one edge $(v_a, v_b)$, (i.e., $(v_a, v_b) \notin \mathcal{M}$ and $(v_a, v_b) \in \mathcal{M}$), then

$$H(\mathcal{M}) - H(\mathcal{M}') = H(S_{ab} + v_a) + H(S_{ab} + v_b)$$
$$- H(S_{ab} + v_a + v_b) - H(S_{ab})$$

where $S_{ab}$ is the minimal separator of $v_a$ and $v_b$ in $\mathcal{M}$.

Thus the change in the entropy of the model after adding (or deleting) an edge is only dependent on the minimal separator of the two vertices in the model graph and as such, this information can be precomputed for all pairs of nodes that are eligible for forward selection. During the forward selection procedure, these "changes" can be associated with the corresponding edge in the clique graph. This also means that after adding a new edge to the model graph, we only have to compute new entropies corresponding to the new edges that are added to the clique graph. In the next theorem, we bound the total number of entropies that have to computed after addition of an edge to the model graph:

**Theorem 4.3** The number of new entropies that need to be computed after adding the edge $(v_a, v_b)$ to the underlying model graph (i.e., after performing one forward selection step) is at most to $2(n - n_a) + 2(n - n_b)$, where $n_a, n_b$ are the number of neighbors of $v_a$ and $v_b$ respectively.

**Proof:** Any new edge that is added to the clique graph must have $C_{ab}$ as one of its endpoints. Consider one such edge, $(C_{ab}, C')$, with associated separator $S' = C_{ab} \cap C'$. This edge must satisfy one of the following properties (cf. Theorem B.3):

- $S' = S_{ab} + v_a$: In this case, the pairs of nodes $(v_x, v_b), v_x \in C' \setminus S'$ are (possibly newly) eligible for forward selection. Each such edge requires computation of $H(S'), H(S' + v_b), H(S' + v_x)$ and $H(S' + v_x + v_b)$. Out of these, $H(S' = S_{ab} + v_a)$ and $H(S' + v_b)$ have already been computed and we only have to compute $H(S' + v_x)$ and $H(S' + v_x + v_b)$.

- $S' \subset S_{ab} + v_a$: The edge $(C', C_a)$ belongs to $CG_{\mathcal{G}'}$ and also to $CG_{\mathcal{G}}$. This implies that pairs of nodes $(v_x, v'_x), v_x \in C' \setminus S', v'_x \in S_{ab} + v_a \setminus S'$ were eligible for addition in $\mathcal{G}$ as well and as such, the entropies needed for these pairs have already been computed. Thus, the only new entropies that need to be computed are for pairs of nodes, $(v_x, v_b), v_x \in C' \setminus S'$ and again we only need to compute $H(S' + v_x)$ and $H(S' + v_x + v_b)$.
- $S' = S_{ab} + v_b$ or $S' \subset S_{ab} + v_b$: The analysis for these two cases is similar.

Thus, the only new entropies that need to be computed are those corresponding to the pairs of nodes of the form $(v_a, v_x)$ or $(v_b, v_x)$ and we need to compute at most two entropies for every such pair. Since there are at most $n - n_a + n - n_b$ such pairs, the total number of new entropies that need to be computed is at most $2(n - n_a) + 2(n - n_b)$. ∎

Note that this theorem assumes that all the entropies required for forward selection in the current model are already computed, and hence it does not apply to the very first step. For example, if we start with the null model (empty model graph), then we need to compute $\binom{n}{2}$ entropies to decide which edge to add in the first step.

## 5 Backward Selection

In this section, we briefly outline how to extend our data structures for doing backward selection as might be required for a procedure which alternates between forward selection and backward selection. Details are deferred to the full version of the paper.

For efficient enumeration of edges eligible for deletion and update of the clique graph data structures, we need to make two changes:

- A matrix of size $n \times n$, indexed by the vertices of the model, is maintained for enumeration of edges eligible for deletion. The entry in the matrix corresponding to a pair of nodes $(u, v)$ tells us whether the edge is present in the model, and if yes, whether it is eligible for deletion.
- The binary index on the separators is augmented to keep the intersection sets for every possible pair of maximal cliques. This information is required to "re-insert" those edges in the clique graph that might have been deleted in the reverse step of adding this edge to the graph.

The algorithms for clique graph update described earlier have to be modified to maintain these data structures as well, but these updates can also be performed in $O(n^2)$ time during both backward selection and forward selection. Also, it can be proved that:

**Theorem 5.1** The number of new entropies that need to be computed after deleting an edge from the underlying model graph (i.e., after performing one backward selection step) is at most $(|S_{ab}| - 1)$ and those are all of the form $H(S_{ab} \setminus \{v_x\})$, for $v_x \in S_{ab}$.

## 6 Mixed Graphs

The results in this paper extend readily to the case of mixed graphs and strong decomposability, via the following theorem:

**Theorem 6.1** [Lau96] Given a strongly decomposable mixed graph $G = (V, E)$, with $V = V_d \cup V_c$, where $V_c$ is the subset of vertices corresponding to the continuous variables and $V_d$ is the subset of vertices corresponding to the discrete variables, then the graph $G' = (V + *, E + E_*)$, with $E_* = \{(*, v) \forall v \in V_d\}$, is chordal.

## 7 Conclusions

Decomposable models possess several important characteristics that make them an appealing class of statistical models, as has been observed in applied contexts ranging from word sense disambiguation [BW94] to multi-dimensional histograms [DGR01]. Efficient algorithms, however, are essential if this class of models is to be exploited in large-scale problems. In this paper, we have presented an efficient new algorithm for performing stepwise selection in decomposable models. The enumeration of edges eligible for forward or backward selection can be a serious practical bottleneck in decomposable models, and the new algorithm is a significant improvement over the naive procedures that are currently used (cf. [Edw95]). Together with methods that allow rapid computation of sufficient statistics, such as the cube computation techniques developed in the database literature [A+96] and techniques that exploit data sparseness [MJ00], we feel that these algorithms allow decomposable models (and their extension to mixtures of decomposable models) to become an increasingly viable alternative in large-scale exploratory data analysis problems.

## Acknowledgements

This work was supported by CONTROL 442427-21389, ONR MURI N00014-00-1-637 and NSF grant IIS-9988642.

## A Proof of Theorem 2.2[4]

The proof of correctness of Theorem 2.2 involves two parts: (i) proving that the graph created after adding such an edge is chordal, (ii) proving that for any chordal graph $\mathcal{G}' = (V, E')$ with $E \subset E'$, $\mathcal{G}'$ can be obtained starting with $\mathcal{G}$ by repeated application of this theorem. Before we proceed with this proof, we will need some definitions:

**Definition:** An *elimination order* on a graph $\mathcal{G}$ is a permutation of its vertices.

**Definition:** Given an elimination order, $v_1, \ldots, v_n$, the *elimination scheme* for this order is as follows: Starting with $v_1$, remove $v_1$ from the graph and add all possible edges between the neighbors of $v_1$. Continue this with $v_2, v_3$ and so on until $v_n$. It can be proved that the graph obtained by adding all these additional edges to $G$ is a triangulated (chordal) graph.

**Definition:** A *perfect elimination order* for a chordal graph is an elimination order which does not add any edges to the graph.

**Lemma A.1** *A graph is* chordal *iff it has a* perfect elimination order.

**Lemma A.2** *The following search algorithm, called* LEXICOGRAPHIC BREADTH-FIRST, *finds a perfect elimination order when applied to a chordal graph.* The algorithm constructs the **reverse** elimination order, starting from the last vertex in the order, which is chosen arbitrarily. At any step, the algorithm chooses a vertex among the remaining vertices that has the highest ordered neighbor. In case of ties, the next highest ordered neighbor is checked and so on (hence the name lexicographic). Any remaining ties are broken arbitrarily. ∎

Given these results, we can proceed with our proofs:

**Theorem A.1** *The graph constructed by adding an edge between two nodes $v_a$ and $v_b$ that satisfy the properties described in Theorem 2.2 for some $S_{ab}$ is chordal.*

**Proof:** We show that there exists an elimination order that can be generated by LEXICOGRAPHIC BREADTH-FIRST algorithm and does not require addition of any edges (and hence, the new graph is chordal). Let the original graph be $\mathcal{G}$, the new graph be $\mathcal{G}'$, and the separator between $v_a$ and $v_b$ be $S_{ab} = s_1, \ldots, s_k$. Note that $S_{ab}$ must be a clique. Also, let the neighbors of $v_b$ in $\mathcal{G}$ be $n_1, \ldots, n_l, s_1, \ldots, s_k$, where $n_1, \ldots, n_l$ are neighbors of $v_b$ not connected to $v_a$.

---

[4]Recently, [GG99] independently derived a characterization for the edges that can be added for forward selection in terms of a junction tree of the model. Their characterization can be shown to be equivalent to our characterization.

Now consider the following partial elimination order for $\mathcal{G}$: $\ldots, v_a, n_1, \ldots, n_l, s_1, \ldots, s_k, v_b$. We claim that this order can be achieved by the LexBFS algorithm, starting with $v_b$, if the ties are broken correctly and hence, it is part of a perfect elimination order. The node just preceding $v_b$ can be any neighbor of $v_b$ and hence we can choose it to be $s_k$. Now note that, $s_{k-1}$ is connected to both $s_k$ and $v_b$ and as such, we can break the next tie in its favor irrespective of the rest of the nodes and so on for $s_{k-2}, \ldots, s_1$. Now, after this is done, the rest of the neighbors of $v_b$ will follow in some order (this order is not relevant to us and without loss of generality, we can assume it to be $n_1, \ldots, n_l$).

This partial elimination order can be replaced by $\ldots, n_1, \ldots, n_l, v_a, s_1, \ldots, s_k, v_b$ without compromising the "perfect-ness" of the elimination order, since it does not add any new edges unless the earlier elimination order is not perfect. Finally, this elimination order is perfect for $\mathcal{G}'$ as well and hence, $\mathcal{G}'$ is chordal. ∎

**Theorem A.2** Given two chordal graphs, $\mathcal{G} = (V, E)$ and $\mathcal{G}' = (V, E')$, with $E \subset E'$, let $T = \mathcal{G}' - \mathcal{G}$, it is possible to reach $\mathcal{G}'$ from $\mathcal{G}$ by repeated application of Theorem 2.2.

**Proof:** This theorem follows from the proof that the *backward selection* procedure is exhaustive [FL89]. ∎

## B Proof of Correctness of the Clique Graph Update Algorithm

For simplicity, we again assume that $C_a$ or $C_b$ are not subsets of $C_{ab}$.

### B.1 Correctness of Step 2:

**Theorem B.1** Given an edge $(C_1, C_2) \in CG_\mathcal{G}$ with separator $S_{12} = C_1 \cap C_2$, if $S_{12} \neq S_{ab}$, then this edge exists in $CG_{\mathcal{G}'}$.

**Proof:** Consider the graph $\mathcal{G} - S_{12}$ (Figure 4). Since $S_{12}$ is a separator for node sets $C_1 \setminus S_{12}$ and $C_2 \setminus S_{12}$, there will be at least two connected components in this graph. Let $P_1$ consist of all the nodes reachable from some node in $C_1 \setminus S_{12}$ and let $P_2$ consist of the nodes reachable from some node in $C_2 \setminus S_{12}$ and let $P_3$ consist of rest of the nodes. So $C_1 \setminus S_{12} \subset P_1$ and $C_2 \setminus S_{12} \subset P_2$. Clearly unless $v_a \in P_1$ and $v_b \in P_2$ or $v_a \in P_2$ and $v_b \in P_1$, the addition of edge $(v_a, v_b)$ does not affect the edge $(C_1, C_2)$.

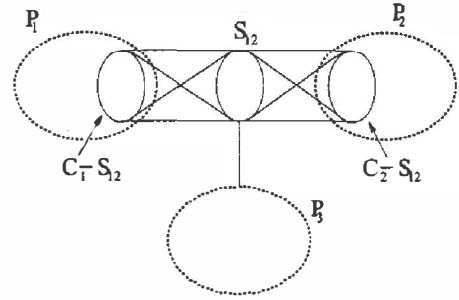

Figure 4: Separation of $\mathcal{G}$ by $S_{12}$

Now, assume $v_a \in P_1$ and $v_b \in P_2$. In that case, $S_{12}$ separates $v_a$ and $v_b$ and hence $S_{ab} \subset S_{12}$ (this follows since $S_{ab}$ is the minimal separator for the nodes $v_a$ and $v_b$). But, since $v_a$ is reachable from $C_1 \setminus S_{12}$ in $\mathcal{G} - S_{12}$ and $v_b$ is reachable from $C_2 \setminus S_{12}$ in $\mathcal{G} - S_{12}$, $S_{ab}$ separates $v_a$ and $v_b$ only if $S_{ab}$ also separates $C_1 \setminus S_{12}$ and $C_2 \setminus S_{12}$. Hence, $S_{12} \subset S_{ab}$. Therefore, $S_{12} = S_{ab}$, which contradicts our assumption. ∎

**Theorem B.2** Given an edge $(C_1, C_2) \in CG_\mathcal{G}$ with separator $S_{12} = C_1 \cap C_2$, if $S_{12} = S_{ab}$, then this edge is not in $CG_{\mathcal{G}'}$ only if $v_a$ is connected to $C_1 \setminus S_{12}$ and $v_b$ is connected to $C_2 \setminus S_{12}$ in $\mathcal{G} - S_{ab}$ or *vice versa*.

**Proof:** Let $P_1$, $P_2$ and $P_3$ be defined as in the preceding proof. This theorem follows from the observation that unless $v_a \in P_1$ and $v_b \in P_2$ (or *vice versa*), the addition of the edge $(v_a, v_b)$ does not have any effect on separability of $C_1 \setminus S_{12}$ and $C_2 \setminus S_{12}$ by $S_{12}$. ∎

### B.2 Correctness of Step 3:

We use the following property of clique graphs in proving correctness of this step:

**Lemma B.1** *[GHP95]* Given a clique graph $CG_\mathcal{G} = (CG_V, CG_E)$, let $C_1, C_2, C_3 \in CG_V$ be such that $(C_1, C_2), (C_2, C_3) \in CG_E$, then:
$$(C_1 \cap C_2) \subset (C_2 \cap C_3) \Rightarrow (C_1, C_3) \in CG_E$$

The following theorem completes the proof of correctness for this step :

**Theorem B.3** If $(C_{ab}, C') \in CG_{\mathcal{G}'}$ and $S' = C_{ab} \cap C'$, then one of the following is true:

1. $S' = S_{ab} + v_a$,
2. $S' = S_{ab} + v_b$,
3. $S' \subset S_{ab} + v_a$ and $(C_a, C') \in CG_{\mathcal{G}'}$ or
4. $S' \subset S_{ab} + v_b$ and $(C_b, C') \in CG_{\mathcal{G}'}$.